\pdfoutput=1

\documentclass[11pt]{article}

\usepackage[final]{acl}

\usepackage{times}
\usepackage{latexsym}

\usepackage[T1]{fontenc}

\usepackage[utf8]{inputenc}

\usepackage{microtype}

\usepackage{inconsolata}

\usepackage{graphicx}
\usepackage{array}
\usepackage{multirow}
\usepackage{booktabs}
\usepackage{colortbl}
\usepackage{xcolor}
\usepackage{vcell}
\usepackage{arydshln}
\usepackage{amsmath}
\usepackage{amsfonts}
\usepackage{tabularx}
\usepackage{makecell}
\usepackage[normalem]{ulem}
\usepackage{CJKutf8}
\usepackage{pifont}
\usepackage{arydshln}
\usepackage{tikz}
\usepackage{pgfplots}
\usepackage{inconsolata}

\usepackage{todonotes}
\usepackage[capitalise,noabbrev]{cleveref}
\usepackage{listings}
\usepackage{fancyvrb} 
\makeatletter
\def\adl@drawiv#1#2#3{%
        \hskip.5\tabcolsep
        \xleaders#3{#2.5\@tempdimb #1{1}#2.5\@tempdimb}%
                #2\z@ plus1fil minus1fil\relax
        \hskip.5\tabcolsep}

\definecolor{verbgray}{gray}{0.9}

\usepackage{colortbl}
\definecolor{lightgray}{rgb}{0.7,0.7,0.7}

\newcommand{\cdashlinelr}[1]{%
  \noalign{\vskip 2pt}   
  \cdashline{#1}[.4pt/2pt] 
  \noalign{\vskip 2pt}   
}
\newcommand{\chinese}[1]{{\begin{CJK*}{UTF8}{gkai} #1 \end{CJK*}}}
\definecolor{light-orange}{HTML}{fee9d4}
\definecolor{light-green}{HTML}{d8f0d3}
\definecolor{light-blue}{HTML}{dae8f5}
\definecolor{light-red}{HTML}{FBC7C4}
\definecolor{set10-red}{HTML}{e41a1c}
\definecolor{set10-blue}{HTML}{377eb8}
\definecolor{set10-green}{HTML}{4daf4a}
\definecolor{bblue}{HTML}{4F81BD}
\definecolor{rred}{HTML}{c4260b}
\definecolor{ggreen}{HTML}{098c1f}
\definecolor{ppurple}{HTML}{9F4C7C}
\definecolor{oorange}{HTML}{F79646}

\usepackage{enumitem}
\setlist[itemize,enumerate]{leftmargin=*}
\usepackage[normalem]{ulem}
\usepackage{pgfplots}
\usetikzlibrary{pgfplots.groupplots}
\pgfplotsset{compat=1.3}
\usepackage{tikz}
\usetikzlibrary{patterns}
\usepackage{xcolor}
\usepackage{todonotes}
\usepackage{listings}
\usepackage{multirow}
\usepackage{graphicx}
\definecolor{CustomBlue}{RGB}{57,83,191}
\usepackage[most]{tcolorbox}

\newtcbox{\clustertab}[1]{on line, box align=base, colback={#1},colframe={#1},size=fbox,arc=2pt,top=-1.5pt, bottom=-1.5pt, left=-1.5pt, right=-1.5pt, boxrule=0pt, enlarge left by=1pt}

\title{MT-RewardTree: A Comprehensive Framework for Advancing LLM-Based Machine Translation via Reward Modeling}

\makeatletter
\def\thanks#1{\protected@xdef\@thanks{\@thanks
        \protect\footnotetext{#1}}}
\makeatother

\author{
    Zhaopeng Feng$^{1\spadesuit}$\quad
    Jiahan Ren$^{1\spadesuit}$ \quad
    Jiayuan Su$^{1\spadesuit}$ \quad 
    Jiamei Zheng$^{1}$ \quad \\
    \bf Hongwei Wang$^{1\dag}$ \quad
    \bf Zuozhu Liu$^{1\dag}$ \\
    $^{1}$Zhejiang University \quad \\
    \texttt{\{zhaopeng.23,jiahan.24,jiayuan.23,jiamei.24\}@intl.zju.edu.cn} \\
    \texttt{\{hongweiwang,zuozhuliu\}@intl.zju.edu.cn} \\
}

\thanks{$^{\spadesuit}$ Equal contribution.} 
\thanks{$^{\dag}$ \space Corresponding author.}

\begin{document}
\maketitle


\begin{abstract}

Process reward models (PRMs) have shown success in complex reasoning tasks for large language models (LLMs). However, their application to machine translation (MT) remains underexplored due to the lack of systematic methodologies and evaluation benchmarks. To address this gap, we introduce \textbf{MT-RewardTree}, a comprehensive framework for constructing, evaluating, and deploying process reward models in MT. Unlike traditional vanilla preference pair construction, we propose a novel method for automatically generating token-level preference pairs using approximate Monte Carlo Tree Search (MCTS), which mitigates the prohibitive cost of human annotation for fine-grained steps. Then, we establish the first MT-specific reward model benchmark and provide a systematic comparison of different reward modeling architectures, revealing that token-level supervision effectively captures fine-grained preferences. Experimental results demonstrate that our MT-PRM-Qwen-2.5-3B achieves state-of-the-art performance in both token-level and sequence-level evaluation given the same input prefix. Furthermore, we showcase practical applications where MT-PRMs successfully identify token-level translation differences and enable test-time alignment for LLMs without additional alignment training. Our work provides valuable insights into the role of reward models in MT research. Our code and data are released in \href{https://sabijun.github.io/MT_RewardTreePage/}{https://sabijun.github.io/MT\_RewardTreePage}.


\end{abstract}

\begin{figure}[ht]
    \centering
    \includegraphics[scale=0.7]{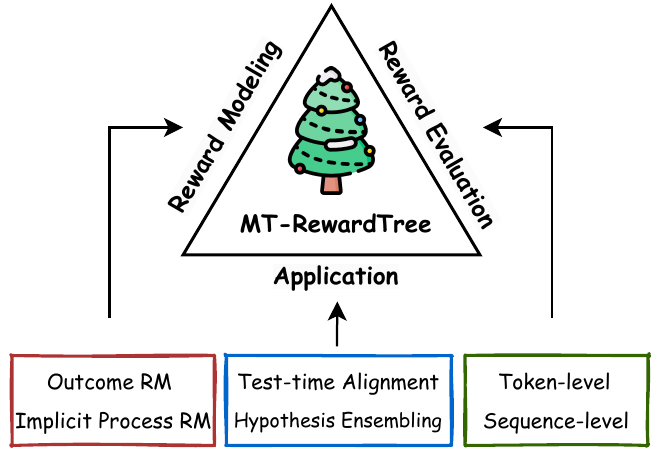}
    \caption{Components of MT-RewardTree.} 
    \vspace{-3mm}
    \label{fig:rounds}
\end{figure}

\section{Introduction}

The next-token prediction process in large language models (LLMs) is often modeled as a Markov Decision Process (MDP) and has achieved remarkable success across various domains, largely attributed to reinforcement learning (RL) and the scaling of test-time compute~\citep{snell2024scaling,zeng2024scaling,deepseekr1,kimi,xiang2025towards}. Reward models are central to these advancements. Outcome Reward Models (ORMs), which are designed to evaluate full responses, have been widely adopted; however, due to the sparsity of outcome rewards, ORMs often yield suboptimal performance and struggle with stability and efficiency during RL training~\citep{lightman,cao-etal-2024-enhancing,chan2024dense}. In contrast, Process Reward Models (PRMs) evaluate intermediate steps to provide fine-grained guidance during both training and inference. PRMs have proven particularly effective in tasks such as mathematics and coding by guiding stepwise decision-making~\citep{wang2024offline,chen2024pad,luo2024improve,qi2024mutual,guan2025rstar}.

Machine translation (MT) naturally aligns with token-level MDP frameworks, as each translation decision corresponds directly to token generation. However, there is still a lack of systematic methodologies for constructing and evaluating PRMs in MT, which has hindered progress relative to advancements in general-domain LLMs.

Developing effective PRMs is challenging. Although \citet{lightman} demonstrate that process supervision with human annotators improves PRM performance in mathematical tasks, this methodology requires domain-expert annotators, resulting in prohibitive costs and practical limitations for translation tasks. Recently, some studies suggest that a PRM can be automatically learned during Direct Preference Optimization (DPO) training~\citep{rafailov2024dpo,rafailov2024from,yuan2024free}. However, existing vanilla preference pair datasets provide only sequence-to-sequence preference data, rather than token-level preferences, which raises concerns about their applicability for token-level alignment. Additionally, evaluating PRMs remains a significant challenge. In mathematical tasks, evaluation is often done using a Best-of-N (BoN) sampling strategy—selecting the highest-scored response from N candidates based on a PRM~\citep{lightman, wang-etal-2024-math, luo2024improve}—or by having the PRM identify errors or verify correctness in the steps~\citep{zheng2024processbench, zhang2025lessons}. Since each step in mathematics has a deterministic answer, these methods do not directly translate to PRM evaluation in MT.

In this paper, we introduce \textbf{MT-RewardTree}, a comprehensive framework for constructing, evaluating, and deploying PRMs in machine translation. We propose an approximate Monte Carlo Tree Search (MCTS) method~\citep{mcts2006, Silver2016MasteringTG} to generate the token-level preference pair dataset. This dataset is then split into a training set for reward model development and a benchmark for reward evaluation. We provide a systematic comparison of different reward modeling methods and test on both token-level and sequence-level performance. Furthermore, we demonstrate two practical applications of PRMs, offering valuable insights for future MT research. Our main contributions are as follows:

\begin{itemize}
    \item We introduce MT-RewardTree, a comprehensive framework for the construction, evaluation, and deployment of PRMs in MT. We establish the first dedicated reward benchmark - MT-PRMBench. Our experiments demonstrate that MT-PRMs achieve competitive performance on both token-level and sequence-level evaluations.
    \item  Comprehensive experiments indicate that our token-level preference pairs, generated through an approximate MCTS method, significantly outperform vanilla preference pairs in process reward model training. Furthermore, our analysis validates that supervising PRMs using preference-based signals is more effective than direct supervision with absolute value estimates.
    \item We demonstrate that our MT-PRMs can directly identify token-level translation differences and facilitate test-time alignment for LLM-based MT without the need for additional alignment training, offering valuable practical insights for the application of reward models in MT.

\end{itemize}

\section{Background}

\subsection{Token-level Markov Decision Process}

LLMs' autoregressive generation can be naturally formulated as a Markov Decision Process, where each token generation is treated as an action. At each time step \(t\), an action \(a_t\) corresponds to the generation of a new token, and the state \(\mathbf{s}_t\) is represented as the sequence of tokens generated up to that point. For tasks that do not involve interaction with an external environment—such as translation—the state is defined as
\[
\mathbf{s}_t = (x_0, \dots, x_L, y_0, \dots, y_{t-1}),
\]
where \((x_0, \dots, x_L)\) represents the input prompt and \((y_0, \dots, y_{t-1})\) is the sequence of generated tokens until time step \(t-1\). The state transition function \(f\) is deterministic and updates the state by concatenating the newly generated token:
\[
\mathbf{s}_{t+1} = f(\mathbf{s}_t, a_t) = \mathbf{s}_t \mid a_t,
\]
with \(\mid\) denoting concatenation.

Within this token-level MDP framework, the reward function \(r(\mathbf{s}_t, a_t)\) is typically designed to provide feedback only at the terminal time step \(T\), reflecting the overall correctness of the generated sequence or the successful completion of the task. To optimize the policy \(\pi_\theta\) based on this reward, Reinforcement Learning with Human Feedback (RLHF)~\citep{ouyang2022training} typically maximizes a KL-constrained objective:
\begin{equation}
\resizebox{0.88\columnwidth}{!}{$
\mathbb{E}_{(s_0,\ldots,s_T) \sim \rho_\pi} \left[ \sum_{t=0}^{T} \left( r(s_t, a_t) - \beta \log\frac{\pi(a_t \mid s_t)}{\pi_{\text{ref}}(a_t \mid s_t)} \right) \right],
$}
\end{equation}
where \(\pi_{\text{ref}}\) is a pre-trained reference policy, \(\beta\) controls the strength of the KL penalty and \(\rho_\pi\) denotes the trajectory distribution induced by policy \(\pi\).

In practice, classical RLHF applies the reward solely at the terminal state. Specifically, the reward function used in Proximal Policy Optimization (PPO)~\citep{schulman2017proximal} is defined as:
\begin{equation}
\resizebox{0.88\columnwidth}{!}{$
r(s_t, a_t) = 
\begin{cases} 
\beta \log \pi_{\text{ref}}(a_t \mid s_t), & \text{if } s_{t+1} \text{ is non-terminal}, \\[1mm]
r(x, y) + \beta \log \pi_{\text{ref}}(a_t \mid s_t), & \text{if } s_{t+1} \text{ is terminal}.
\end{cases}
$}
\end{equation}

\subsection{Reward Modeling in RLHF}

Reward modeling is the cornerstone of RLHF, enabling LLMs to align their outputs with human preferences. In this section, we distinguish between typical (sequence-level) reward modeling and the more fine-grained token-level approach.

\noindent \textbf{Sequence-level Reward Modeling.}  
In classical RLHF, the reward function is learned from human feedback on prompt-response pairs \((\mathbf{x}, \mathbf{y})\). The reward model is formulated as a contextual bandit, where a scalar reward is assigned only at the terminal state—i.e., once the full response sequence has been generated. This formulation, known as Outcome Reward Modeling, follows the Bradley-Terry~\citep{Bradley1952RANKAO} preference model to define the probability of preferring one response over another:
\begin{equation}
\resizebox{0.8\columnwidth}{!}{$
p^*(\mathbf{y}^w \succeq \mathbf{y}^l) = \frac{\exp\left(r_\phi(\mathbf{x}, \mathbf{y}^w)\right)}{\exp\left(r_\phi(\mathbf{x}, \mathbf{y}^w)\right) + \exp\left(r_\phi(\mathbf{x}, \mathbf{y}^l)\right)}.
$}
\end{equation}

To train the reward model \( r_\phi \), we construct a preference dataset \( \mathcal{D} \), where each prompt \( x \) is paired with two candidate responses, \( y \) and \( y' \). Human annotators or heuristics determine the preferred response \( y_w \) and the rejected response \( y_l \). The reward model is then optimized to maximize the likelihood of these human preferences:
\begin{equation}
\label{ormeq}
\resizebox{0.85\columnwidth}{!}{$
\max_\phi \mathbb{E}_{(x, y_w, y_l) \sim \mathcal{D}} \left[ \log \sigma(r_\phi(x, y_w) - r_\phi(x, y_l)) \right],
$}
\end{equation}  
where \( \sigma \) is the logistic function. By training \( r_\phi \) in this manner, we ensure that the model assigns higher rewards to preferred responses, effectively capturing human-like quality judgments for sequence-level evaluation.

\begin{figure*}[ht]
    \centering
    \includegraphics[scale=0.7]{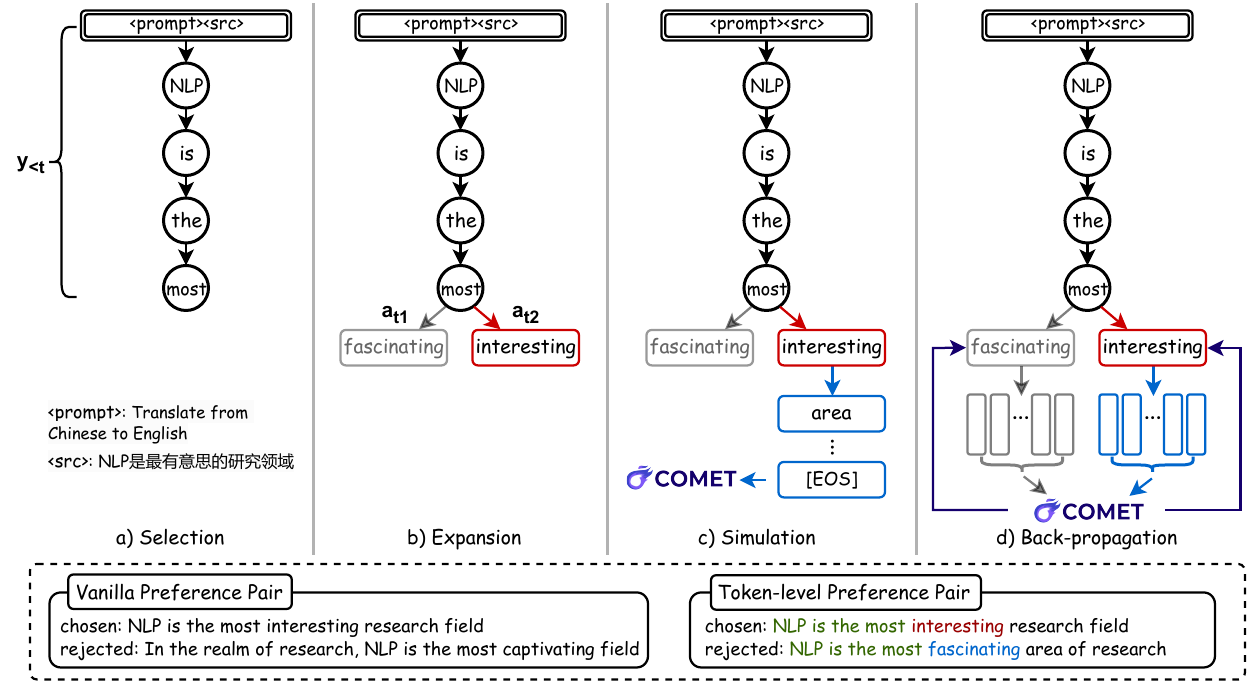}
    \caption{The construction process of token-level preference pairs. We utilize TowerInstruct-7B-v0.2 to generate candidate tokens. A \textit{token-level} preference pair comprises two translations that share an identical prefix. } 
    \vspace{-3mm}
    \label{fig:mcts}
\end{figure*}

\noindent\textbf{Token-level Reward Modeling.}  
While the sequence-level approach treats the entire generated response as a single action, it fails to capture the fine-grained decision-making process inherent in token generation. Token-level reward modeling addresses this limitation by evaluating rewards at each token-generation step. This approach corresponds to a form of Process Reward Models. The cumulative reward for a trajectory \(\tau\) is computed as the sum of per-token rewards, and the corresponding preference probability between two trajectories, \(\tau^w\) and \(\tau^l\), is given by:
\begin{equation}
\resizebox{0.85\columnwidth}{!}{$
p^*(\tau^w \succeq \tau^l)=\frac{\exp\left(\sum_{i=1}^N r(\mathbf{s}_i^w, a_i^w)\right)}{\exp\left(\sum_{i=1}^N r(\mathbf{s}_i^w, a_i^w)\right)+ \exp\left(\sum_{i=1}^M r(\mathbf{s}_i^l, a_i^l)\right)}.
$}
\end{equation}

Although token-level reward modeling offers finer-grained feedback, obtaining effective PRMs is more challenging to obtain and deploy~\citep{lightman,cao-etal-2024-enhancing}.

\section{MT-RewardTree}

In this section, we introduce the components of the MT-RewardTree. We first describe how we construct token-level preference pairs using an MCTS-based method. Next, we review several approaches employed for reward modeling.


\subsection{Constructing Token-level Preference Pairs}
Prior studies have investigated translation preference pair construction~\citep{xu2024contrastive,mt-pref,feng-etal-2024-ladder}, yet a standardized token-level preference pair dataset for PRMs in MT remains absent. MQM~\citep{freitag-etal-2021-experts} datasets depend on manual error annotation, which is both cost-prohibitive and incapable of producing granular token-level preference pairs.

Drawing inspiration from MCTS, we propose a token-centric approach that quantifies token quality based on its potential to contribute to higher-quality translations. This method aligns with Monte Carlo-based PRMs construction techniques in mathematics, where step-wise quality is determined by its incremental contribution to deriving correct answers~\citep{wang-etal-2024-math,guan2025rstar}.

The MCTS process consists of four main steps (depicted in Figure~\ref{fig:mcts}): Selection, Expansion, Simulation (Evaluation), and Back-propagation.

\begin{enumerate}
\item \textbf{Selection}: The first phase involves selecting a portion of the existing tree that is most promising for further expansion. Starting from the root node, a standard approach would traverse the tree down to a leaf using the PUCT algorithm \citep{Rosin2011, Silver2017}. Since our goal is to construct token-level preference pairs rather than achieving global optimality, we automatically select the existing prompt and previously generated tokens as the prefix \( y_{<t} \).

\item \textbf{Expansion}: If the selected leaf node is not an EOS (end-of-sentence) token—i.e. if it is not a terminal state—the node is expanded by generating \( k \) candidate children. This is achieved by decoding one additional step using the language model and selecting the top-\( k \) tokens as the new children. We select the top-2 candidate tokens \( a_{tj} \) (with \( j \in \{1,2\} \)) that have the highest logits. Preliminary experiments demonstrate that tokens outside of the top-2 yield significantly lower translation quality during the \textbf{Simulation} phase. These top-2 tokens, sharing the same prefix \( y_{<t} \), form the basis for our token-level preference pair.

\item \textbf{Simulation (Evaluation)}: From each expanded node \( a \), we generate \(n\) complete translation rollouts until an EOS token is reached. We then evaluate the quality (or groundedness) of the full translation sequence, denoted by \( g(y, n) \).  In our framework, we use COMETKiwi~\citep{rei2022cometkiwi} to estimate the quality of all \( n \) full rollouts. These scores are averaged and further assigned as the value of node \( a \), i.e., \( V(a) \). 

\item \textbf{Back-propagation}:
Since our objective is to construct token-level preference pairs, we compare the values \( V(a_{t1}) \) and \( V(a_{t2}) \) to determine which expanded token is superior. Finally, we retain the node with the higher \( V \) value. This node, along with its corresponding prefix \( y_{<t} \), is then used as the starting point in the next simulation cycle, beginning again at Step 1.

\end{enumerate}

These four steps are repeated until the EOS token appears during the Selection phase. We retain one rollout from the superior token and one from the inferior to construct our token-level preference pair. We use COMETKiwi to guarantee the score gap lies between 0.04 and 0.4 to control the quality.

\begin{table*}[t]
    \centering
    \small
    \setlength{\tabcolsep}{3pt}
    \resizebox{0.85\textwidth}{!}{
    \begin{tabular}{lccccccc}
        \toprule[0.5mm]
        \multirow{2}{*}{Model} & \multicolumn{7}{c}{MT-PRMBench} \\
        & \multicolumn{3}{c}{Sequence-level} && \multicolumn{3}{c}{Token-level}  \\
        \cline{2-4} \cline{6-8} 
        \\
        & EN→XX & XX→EN & Avg. && EN→XX & XX→EN & Avg. \\
        \midrule

        \multicolumn{4}{l}{\textcolor{gray}{\textit{Baselines}}}\\

        Skywork-Reward-LLaMA-3.1-8B & 0.857 & 0.773 & 0.815 && - & - & - \\
        MT-Ranker-base & 0.785 & 0.787 & 0.786 && - & - & - \\
        MT-Ranker-large & 0.847 & 0.873 & 0.860 && - & - & -\\
        \cdashlinelr{1-8}
        \multicolumn{4}{l}{\textcolor{gray}{\textit{PRMs}}}\\
        MT-PRM-LLaMA-3.2-3B & 0.777 & 0.775 & 0.776 && 0.542 &  0.615 & 0.578 \\
        MT-PRM-Qwen-2.5-3B & 0.867 & 0.858 & 0.863 && 0.637 & 0.685 & 0.660 \\

        \bottomrule[0.5mm]
    \end{tabular}
    }
    \caption{Accuracy results on MT-PRMBench. Skywork-Reward-LLaMA-3.1-8B is an advanced ORM for general domains, while MT-Ranker represents the SoTA non-metric reference-free translation quality estimation model.}
    \label{tab:main_results}
    \vspace{-3mm}
\end{table*}

\begin{table}[ht]
\centering
\small
\renewcommand\arraystretch{1.25}
\begin{tabular}{cccc}
\toprule
\multirow{2}{*}{Translation Direction}  & \multicolumn{2}{c}{Token-level Preference Pairs} \\
\cline{2-3}
 & Train & MT-PRMBench \\ 
\midrule
DE-EN & 1,255 & 200 \\
EN-DE & 2,059 & 200  \\
RU-EN & 1,219 & 200 \\
EN-RU & 1,711 & 200  \\
ZH-EN & 1,232 & 200 \\
EN-ZH & 1,176 & 200  \\
\bottomrule
\end{tabular}
\caption{Data Statistics.}
\label{tab:datasets}
\end{table}

\subsection{Implicit Process Reward Modeling}



Unlike ORMs, which assign a single reward to the entire response, PRMs aim to assign rewards at a finer granularity, such as at each step or token. However, traditional PRMs training requires step-level annotations, which are costly to obtain. Recent studies~\citep{rafailov2024from, zhong2024dpo} show that ORMs can be trained with implicit reward modeling, enabling PRMs to emerge naturally without the need for explicit step labels.

Consider an ORM where the reward is parameterized by the log-likelihood ratio of two causal language models:
\begin{equation}
r_\theta(\mathbf{y}) := \beta \log \frac{\pi_\theta(\mathbf{y})}{\pi_\text{ref}(\mathbf{y})}
\end{equation}
where \( \pi_\theta \) represents the trained model’s probability distribution, and \( \pi_\text{ref} \) is a reference model. We define the cumulative reward up to step \( t \) as:
\begin{equation}
q_\theta^t(\mathbf{y}_{<t}, y_t) := \sum_{i=1}^{t} \beta \log \frac{\pi_\theta(y_i | \mathbf{y}_{<i})}{\pi_\text{ref}(y_i | \mathbf{y}_{<i})}
\end{equation}
which serves as an exponential moving average of \( r_\theta \) across steps. The expected process reward at step \( t \) can then be expressed as:
\begin{equation}
q_\theta^t(\mathbf{y}_{<t}, y_t) = \beta \log \mathbb{E}_{\pi_\text{ref}(\mathbf{y}|\mathbf{y}_{\leq t})} \left[ e^{\frac{1}{\beta} r_\theta(\mathbf{y})} \right]
\end{equation}
This formulation shows that \( q_\theta^t \) is an exact expectation of the outcome reward \( r_\theta \) at step \( t \), making it analogous to a Q-value in reinforcement learning.

By defining the process reward \( r_\theta^t \) as the difference between successive Q-values:
\begin{equation}
\label{eq:9}
r_\theta^t := q_\theta^t - q_\theta^{t-1} = \beta \log \frac{\pi_\theta(y_t | \mathbf{y}_{<t})}{\pi_\text{ref}(y_t | \mathbf{y}_{<t})}
\end{equation}

We see that PRMs can be derived directly from an ORM trained on response-level data, without requiring explicit step-wise labels. This insight suggests that training an ORM inherently leads to the learning of a Q-function, enabling step- or token-level reward modeling without requiring additional supervision. A typical example of this is DPO~\citep{rafailov2024dpo}, which optimizes the following objective:  
\begin{equation}
\resizebox{1.0\columnwidth}{!}{$
L_{\text{DPO}}(\pi_\theta; \pi_{\text{ref}}) = -\mathbb{E}_{(x, y_w, y_l) \sim \mathcal{D}} \log \sigma \left( \beta \log \frac{\pi_\theta(y_w | x)}{\pi_{\text{ref}}(y_w | x)} - \beta \log \frac{\pi_\theta(y_l | x)}{\pi_{\text{ref}}(y_l | x)} \right)
$}
\end{equation}
This formulation shows that optimizing \( \pi_\theta \) implicitly optimizes a reward model, as described in Eq.~\ref{ormeq}. Moreover, \citet{yuan2024free} demonstrated that this approach is agnostic to the specific training objective (i.e., not limited to DPO). It can be instantiated using various training objectives (e.g., KTO~\citep{ethayarajh2024kto}), with the only modification being the substitution of \( r_\theta(\mathbf{y}) \) with \( \beta \log \frac{\pi_\theta(\mathbf{y})}{\pi_\text{ref}(\mathbf{y})} \).  

Moreover, our implicit PRMs can seamlessly be converted into ORMs using weighted implicit rewards:
\begin{equation}
\label{eq:weighted dpo}
    r_{\text{sequence}}(y_{1:T}) = \sum_{k=0}^{T-1} w_t \log\frac{\pi_\theta(y_t|y_{<t})}{\pi_\text{ref}(y_t|y_{<t})}
\end{equation}
where the positional weights \( w_t = \frac{1}{|y_{<t}|} \) are used to balance the contributions of each token.



\begin{table}[ht]
\centering

\resizebox{0.9\columnwidth}{!}{%
\begin{tabular}{lcccccccccc}
\toprule[0.5mm]

 \textbf{Preference Pair Type}  & \textbf{Training Strategy} & \textbf{Avg.} \\\midrule
 \raggedleft{Token-level} & {DPO} & {0.660}\\
 \raggedleft{Vanilla} & {DPO} & {0.574}\\
  \raggedleft{Token-level} & {KTO} & {0.644}\\
 \raggedleft{Vanilla} & {KTO} & {0.562} \\
\bottomrule[0.5mm]
\end{tabular}
}
\caption{Ablation study on the effect of training preference data and implicit reward training objectives. The backbone model is Qwen-2.5-3B-Instruct and we test these variants on MT-PRMBench (Token-level).}
\vspace{-3mm}
\label{tab:ablation}
\end{table}

\section{Experiments}
\subsection{Experimental Setup}

\noindent{{\textbf{Datasets.}}}
We explore four languages—English (EN), German (DE), Chinese (ZH), and Russian (RU)—and six translation directions: EN→XX and XX→EN. Our raw corpus consists of test sets from WMT17 to WMT20, supplemented with development and test sets from the Flores~\citep{costa2022no}. We use the TowerInstruct-7B-v0.2\footnote{\href{https://huggingface.co/Unbabel/TowerInstruct-7B-v0.2}{https://huggingface.co/Unbabel/TowerInstruct-7B-v0.2}} model with a temperature of 0.95 and apply the MCTS-based approach described earlier. During the Simulation step, we sample three candidate hypotheses for each node. Our token-level preference pairs is divided into train and test set (MT-PRMBench). Detailed statistics are in Table~\ref{tab:datasets}.


\noindent \textbf{Training Details.}
We take LLaMA-3.2-3B-Instruct and Qwen-2.5-3B-Instruct as the backbone models for training. For the DPO training, the higher-scored sentence is designated as the chosen response, while the lower-scored sentence is labeled as the rejected response. For the KTO training, the higher-scored sentence is treated as the positive sample, and the lower-scored sentence as the negative sample. We set $\beta$ as 0.1.

\noindent{\textbf{Reward Evaluation.}}  
We evaluate reward models by framing the task as a classification problem, similar to prior work on reward model benchmarks in the general domain~\citep{lambert2024rewardbench, liu2024rm}. For sequence-level evaluation, given a tuple \( (x, y_c, y_r) \), where \( x \) is the prompt, \( y_c \) is the chosen response, and \( y_r \) is the rejected response, the reward model predicts whether \( y_c \) is better than \( y_r \). If the reward model assigns a higher reward to \( y_c \) than to \( y_r \), the prediction is correct; otherwise, it is incorrect. We use accuracy as the evaluation metric, computed as follows:
\begin{equation}
\label{eq:acc}
\resizebox{1.0\columnwidth}{!}{$
\text{Accuracy} = \frac{1}{|D|} \sum_{(x, y_c, y_r) \in D} I[R_\theta(x, y_c) > R_\theta(x, y_r)]
$}
\end{equation}
where \( I(\cdot) \) is the indicator function, and \( D \) denotes the evaluation dataset.

For token-level evaluation, we use tuples of the form \( (x, y_{<t} + a_c, y_{<t} + a_r) \), where \( y_{<t} \) is the generated tokens before, \( a_c \) is the next chosen token, and \( a_r \) is the rejected token. Similarly, we compute accuracy as the evaluation score: if the PRM assigns a higher reward to \( a_c \) than to \( a_r \), the prediction is correct; otherwise, it is incorrect.

\noindent{{\textbf{MT-PRMBench.}}} MT-PRMBench comprises two distinct subsets for evaluation: \textit{Token-level} and \textit{Sequence-level}. The \textit{Token-level} subset is designed for assessing preferences between immediate next-token candidates that follow an identical input prefix. In contrast, the \textit{Sequence-level} facilitates the comparison of entire generated sequence completions that also originate from a shared input prefix.

\subsection{Evaluation Results}
\label{sec:main_eval}
\noindent{\textbf{Token-level Performance.}}
From Table~\ref{tab:main_results}, we can observe that our MT-PRM-LLaMA-3.2-3B and MT-PRM-Qwen-2.5-3B models achieved accuracies of 0.578 and 0.66 respectively on the token-level MT-PRMBench. As shown in Table~\ref{tab:ablation}, we systematically compare models trained with vanilla sequence-level preference pairs versus our token-level preference pairs, while evaluating both DPO and KTO training objectives. The results demonstrate that token-level preference pairs significantly improve discrimination accuracy: implicit PRMs trained with token-level preference pairs outperform vanilla sequence-level baselines by +8.6\% (DPO) and +11.5\% (KTO). This performance gap highlights the critical advantage of token-level preference pairs in helping capture fine-grained translation quality distinctions.

\noindent{\textbf{Sequence-level Performance.}} We also convert our PRMs to sequence-level scoring through weighted DPO rewards (as shown in Eq.~\ref{eq:weighted dpo}). We can observe that our MT-PRM-Qwen-2.5-3B achieves the highest performance among all models in the Prefixed set, with an average score of 0.863, outperforming both Skywork-Reward-LLaMA-3.1-8B~\footnote{\href{https://huggingface.co/Skywork/Skywork-Reward-Llama-3.1-8B}{https://huggingface.co/Skywork/Skywork-Reward-Llama-3.1-8B}} and the MT-Ranker~\citep{moosa2024mt} variants. This demonstrates the effectiveness of our token-level supervision framework even when adapted to sequence-level scoring.

\begin{table*}[t]
    \centering
    \small
    \resizebox{0.8\textwidth}{!}{%
    \begin{tabular}{lccccccc}
    \toprule

    \textbf{MT-PRMBench (Token-level)} & \textbf{DE-EN} & \textbf{RU-EN} & \textbf{ZH-EN} & \textbf{EN-DE} & \textbf{EN-RU} & \textbf{EN-ZH} & \textbf{Avg.} \\
    \midrule
    Supervised by Value (SV) & 0.52 & 0.56 & 0.48 & 0.50 & 0.51 & 0.55 & 0.52 \\
    Supervised by Preference (SP) & 0.64 & 0.74 & 0.68 & 0.58 & 0.68 & 0.66 & 0.66 \\
    \bottomrule
        
    \end{tabular}}
    \caption{\label{tab:value_preference} Experimental results of two training methods on the Token-level Benchmark}
    \label{tab:supervise}
    \vspace{-3mm}
    \end{table*}

\begin{table*}[t]
    \centering
    \small
    \resizebox{\textwidth}{!}{%
    \begin{tabular}{lccccccccccc}
    \toprule
    \multicolumn{1}{c}{Source} & \multicolumn{11}{l}{\chinese{在橘红色背景墙映衬下格外鲜亮的广告牌上，它们各自标出了不菲的价码。}} \\
    \cdashlinelr{1-12}
    \multicolumn{1}{c}{Reference} &  \multicolumn{11}{l}{On the bright advertising board against a tangerine colored wall, they listed their respective exorbitant price tags.}\\
    \cdashlinelr{1-12}
    \multicolumn{1}{c}{Translation A} &  \multicolumn{11}{l}{Against an orange background, the \textbf{bright} billboards listed their respective, \textbf{hefty} prices.}\\

    \multicolumn{1}{c}{Translation B} &  \multicolumn{11}{l}{Against an orange background, the \textbf{dark} billboards listed their respective, \textbf{hefty} prices.}\\

    \multicolumn{1}{c}{Translation C} &  \multicolumn{11}{l}{Against an orange background, the \textbf{bright} billboards listed their respective, \textbf{bargain} prices.}\\
    \midrule
    \multicolumn{1}{c}{Translation A} & 'Against' & ' an' & … & ' the' & \textbf{' bright'} & ' bill' & … & \textbf{' hefty'} & ' prices' & \textbf{Weighted Implicit Rewards(↑)} & \textbf{COMETKiwi(↑)}\\
    \multicolumn{1}{c}{Reward} & -2.3 & -2.3 & … & 0.07 & 1.28 & -2.3 & … & 0.03 & 0.09 & -3.43 & 0.80\\
    \cdashlinelr{1-12}
    \multicolumn{1}{c}{Translation B} & 'Against' & ' an' & … & ' the' & {\color{red}{\textbf{' dark'}}} & ' bill' & … & ' hefty' & ' prices' & \textbf{Weighted Implicit Rewards(↑)} & \textbf{COMETKiwi(↑)}\\
    \multicolumn{1}{c}{Reward} & -2.3 & -2.3 & … & 0.07 & -2.3 & -2.3 & … & 0.09 & 0.09 & -4.13 & 0.73\\
    \cdashlinelr{1-12}
    \multicolumn{1}{c}{Translation C} & 'Against' & ' an' & … & ' the' & ' bright' & ' bill' & … & \color{red}{\textbf{' bargain'}} & ' prices' & \textbf{Weighted Implicit Rewards(↑)} & \textbf{COMETKiwi(↑)}\\
    \multicolumn{1}{c}{Reward} & -2.3 & -2.3 & … & 0.07 & 1.28 & -2.3 & … & -0.95 & -2.3 & -3.99 & 0.78\\
    \midrule

    \end{tabular}}
    \caption{\label{tab:case_study_credit} Case study illustrating token-level credit assignment by our Qwen-PRM.}
    \vspace{-3mm}
    \end{table*}

\section{Analysis and Practical Insight}
\subsection{Modeling Advantage versus Value as the PRM Training Signal}
We have explored the impact of vanilla preference pairs versus MCTS-generated token-level preference pairs on the performance of implicit PRMs in the previous experiments (Table~\ref{tab:ablation}). This section shifts focus to the nature of the supervisory signal used for training PRMs. Specifically, we analyze our choice of implicit reward modeling—derived from token-level preference pairs (termed "Supervised by Preference (SP)")—against the alternative of directly using MCTS-backpropagated values $V(a)$ as a dense supervisory signal (termed "Supervised by Value (SV)").

Table~\ref{tab:supervise} demonstrates that SP yields markedly superior performance in token-level discrimination, where SP achieved an average accuracy of 0.66, while SV achieved 0.52. This significantly higher accuracy underscores the effectiveness of our implicit PRM when trained with preference-based supervision. The SV approach, directly regressing on MCTS-backpropagated $V(a)$ values, tasks the PRM with learning a value function. However, $V(a)$ as a dense, token-level reward faces limitations: Monte Carlo estimates can be noisy, and the absolute value of a partial translation may offer an indirect and insufficiently discriminative signal for the most recent token's quality, especially in complex MT scenarios or with imperfect rollouts—a recognized challenge in process supervision literature~\citep{lightman}. We also find that these $V(a)$ values often clustered within a narrow numerical range (e.g., 0.7-0.8) further illustrates why this SV approach struggles. Such clustering severely hampers a regression model’s ability to discern fine-grained distinctions, as the supervisory signal becomes subtle.

In contrast, the SP approach, using DPO/KTO objectives, excels by directly learning a representation reflecting the \textit{advantage} of one token choice over another. This implicitly shapes a reward function where the per-step process reward, $r_{\theta}^{t} = \beta\log\frac{\pi_{\theta}(y_{t}|y_{<t})}{\pi_{ref}(y_{t}|y_{<t})}$ (Eq.~\ref{eq:9}), quantifies the localized, step-wise \textit{advantage} of selecting token $y_t$. This preference-based optimization strategy is well-supported by prior work~\citep{yuan2024free, rafailov2024from, wang2024offline}. Crucially, DPO’s mechanism, focusing on the log-probability ratio between sequences (Eq.~\ref{ormeq}), is inherently more sensitive to relative differences than absolute value regression. This makes it adept at capturing preferences even when underlying $V(a)$ scores from the SV context are close, explaining its superior performance in training PRMs for nuanced token-level discrimination in our MT setting. 

\subsection{Per-token Credit Assignment}
Our PRM can identify token-level translation differences, with the per-step process reward defined as Eq.~\ref{eq:9} quantifying the reward for generating token $y_t$ at step $t$. These individual token rewards are then aggregated into a final weighted sequence score (Eq.~\ref{eq:weighted dpo}), allowing for a comprehensive evaluation that originates from fine-grained assessments.

Table~\ref{tab:case_study_credit} provides a case study illustrating these capabilities. For instance, in Translation B, the token "dark", which semantically contradicts the source "\chinese{鲜亮}" (bright), receives a significantly negative reward (-2.3) from our PRM. In Translation C, the incorrect token "bargain", used where "\chinese{不菲的}" (exorbitant/costly) is implied, is substantially penalized (-0.95). In contrast, contextually appropriate tokens in Translation A, such as "bright" (1.28) and "hefty" (0.03), secure relatively higher rewards. Furthermore, the final weighted sequence scores computed by our PRM demonstrate strong alignment with automatic metrics like COMETKiwi. Translation A, the highest quality hypothesis (COMETKiwi 0.80), also achieves the most favorable weighted PRM score (-3.43). This case study thus substantiates our PRM's effective token-level credit assignment and the consistency of its fine-grained assessments with established sequence-level quality metrics.

\subsection{Test-time Alignment}

Test-time alignment, also known as decoding-time alignment~\citep{huang2024deal,rashid2024critical}, refers to the process of adjusting an LM's output during inference to better align with human preferences, without additional training or fine-tuning. Its application in MT remains underexplored.

In the context of MT, given the prior context \( s_{<t} \) and timestamp \( t \), we define the reward-guided scoring function for a candidate token \( a \) as:
\begin{equation}
\label{eq:tta}
\
s(a, s_{<t}) = \text{LM}(a \mid s_{<t}) + w \cdot P(r([s_{<t}, a]))
\
\end{equation}
where \( \text{LM}(a \mid s_{<t}) \) represents the LM’s predicted probability for token \( a \) given the preceding context \( s_{<t} \). \( r([s_{<t}, a]) \) denotes the reward signal for token \( a \), conditioned on the prior context \( s_{<t} \). The softmax function is applied over the reward signal \( r([s_{<t}, a]) \), computed over the top \( k \) candidate tokens (with \( k \) being a window size), normalizing the reward value, which we label as \(P(r([s_{<t}\), a]).  The scaling factor \( w \) adjusts the relative weight of the reward signal, allowing it to contribute effectively without overpowering the LM's probability. Compared to standard decoding strategies, this approach offers a more refined scoring function, as it encourages the generated text to: 1) Maintain semantic coherence and relevance with the prior context, and 2) Align more closely with reward-based criteria and human preferences. Test-time alignment also substantially reduces the need for the extensive resources typically required for LM alignment training.

\begin{figure}[h]
    \centering
    \includegraphics[scale=0.55]{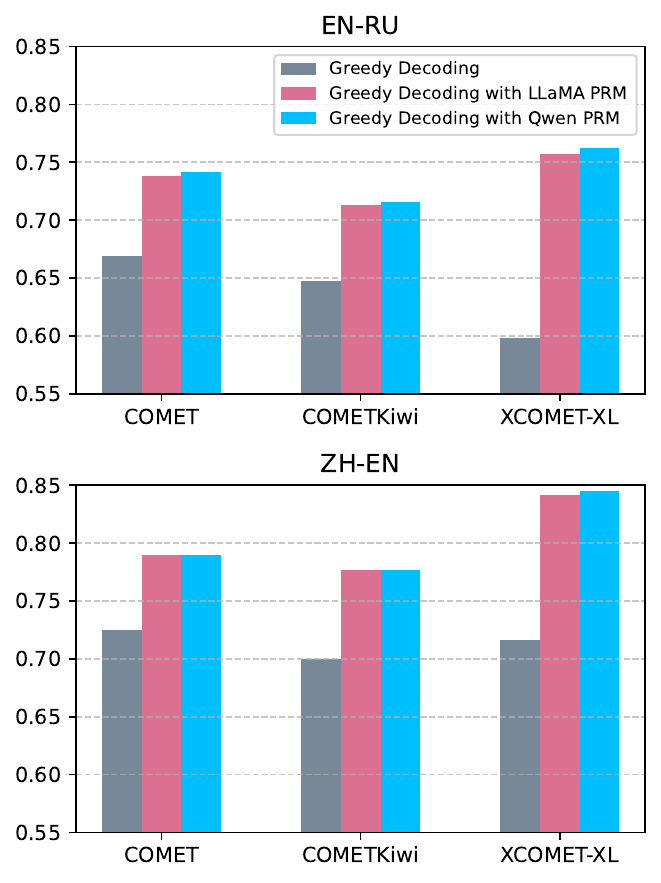}
    \vspace{-2mm}
    \caption{Results of test-time alignment across WMT 23 ZH-EN and EN-RU. MT-PRMs with less parameters can assist in aligning Qwen-2.5-14B-Instruct.} 
    \vspace{-3mm}
    \label{fig:zhen}
\end{figure}

We use Qwen2.5-14B-Instruct\footnote{\href{https://huggingface.co/Qwen/Qwen2.5-14B-Instruct}{https://huggingface.co/Qwen/Qwen2.5-14B-Instruct}} for generating tokens and leverage MT-PRM-LLaMA-3.2-3B and MT-PRM-Qwen-2.5-3B as the models for providing token-level rewards. We randomly sample 500 cases from the WMT 2023 testset. As shown in Figure~\ref{fig:zhen}, the reward-guided decoding methods outperform the standard greedy decoding in both EN-RU and ZH-EN translation tasks, evaluated by the COMET~\citep{rei2020comet}, COMETKiwi~\citep{rei2022cometkiwi}, and XCOMET-XL~\citep{guerreiro2024xcomet} metrics. For instance, using the XCOMET-XL metric, LLaMA PRM and Qwen PRM outperform the standard greedy decoding by 17.5\% and 17.9\% in the EN-RU task respectively. Additionally, Qwen PRM slightly outperforms LLaMA PRM in both translation tasks and across all metrics, which aligns with the results in Table~\ref{tab:main_results}, where Qwen PRM achieves better token-level reward performance. These findings highlight the effectiveness of reward-guided decoding strategies in improving MT outcomes.

A crucial aspect of reward-guided alignment is the trade-off with output diversity. Our framework explicitly manages this balance via the weighting parameter \( w \) in the reward-guided scoring function (Eq.~\ref{eq:tta}). The parameter \( w \) controls the influence of the PRM: a higher \( w \) prioritizes the learned preferences for stronger alignment, while a lower \( w \) preserves more of the base model's original output distribution, thereby maintaining diversity.


\begin{table}[h]
\centering
\small
\resizebox{\columnwidth}{!}{%
\begin{tabular}{llcccc}
\toprule
\textbf{Model} & \textbf{Task} & \textbf{w=0} & \textbf{w=0.3} & \textbf{w=0.5} & \textbf{w=0.7} \\
\midrule
\multirow{2}{*}{\textit{MT-PRM-Llama Guided}} & ZH-EN & 0.7167 & 0.8759 & 0.8420 & 0.8677 \\
& EN-RU & 0.5984 & 0.7459 & \textbf{0.7619} & 0.7143 \\
\midrule
\multirow{2}{*}{\textit{MT-PRM-Qwen Guided}} & ZH-EN & 0.7167 & \textbf{0.8723} & 0.8448 & 0.8671 \\
& EN-RU & 0.5984 & 0.7444 & 0.7569 & 0.6721 \\
\bottomrule
\end{tabular}
}
\caption{XCOMET scores showing the impact of the weighting parameter \(w\) on test-time alignment. The baseline is Qwen2.5-14B-Instruct (\(w=0\)). Optimal values vary, allowing for tunable performance.}
\label{tab:tradeoff_w}
\end{table}

The quantitative results in Table~\ref{tab:tradeoff_w} show that a range of \(w\) values provides a significant boost over the baseline (\(w=0\)), allowing users to choose a balance that suits their needs. For example, with the Llama-PRM on EN-RU, \(w=0.5\) yields the best XCOMET score, while \(w=0.3\) is optimal for the MT-PRM-Qwen. Furthermore, the qualitative case study in Table~\ref{tab:diversity_case} illustrates that even under PRM guidance, different models and weights produce varied, high-quality outputs, demonstrating that diversity can be maintained.

\begin{table}[h]
\centering
\small
\begin{tabular}{p{0.3\linewidth} p{0.6\linewidth}}
\toprule
\textbf{System} & \textbf{Translation} \\
\midrule
\textbf{Source (Reference)} & \chinese{放入充电盒中，耳机自动关机。} (Put it in the charging case, the headphones will turn off automatically.) \\
\midrule
Qwen2.5-14B-Instruct  (\( w = 0 \)) & Put them in the charging case, and the headphones will shut off automatically. \\
MT-PRM-Llama (\( w = 0.3 \)) & Put them in the charging case, and the headphones will shut off automatically. \\
MT-PRM-Qwen (\( w = 0.5\)) & When placed in the charging case, the headphones shut off automatically. \\
MT-PRM-Llama (\( w = 0.7 \)) & Place the earbuds in the charging case, and they will shut down automatically. \\
\bottomrule
\end{tabular}
\caption{Qualitative case study illustrating output diversity under PRM guidance.}
\label{tab:diversity_case}
\end{table}

\section{Related Work}

\noindent\textbf{Token-Level Feedback Mechanisms.}  
Fine-grained feedback has been recognized for its ability to help models capture potential errors more precisely~\citep{lightman}. In the context of mathematical reasoning, process supervision using Monte Carlo methods has shown significant promise~\citep{wang-etal-2024-math, qi2024mutual,guan2025rstar}. Furthermore, developments in general-domain have demonstrated that DPO can implicitly learn token-level rewards through policy optimization, a process referred to implicit reward learning~\citep{rafailov2024from, wang2024offline, yuan2024free}. Despite these advancements, these approaches have yet to be tested in the context of MT. The translation community has long acknowledged the value of granular feedback, with early attempts relying on binary error markings from human annotations~\citep{kreutzer2020correct}, reference-based heuristics~\citep{petrushkov-etal-2018-learning}, or LLM~\citep{feng2024improving}. 

\noindent\textbf{Alignment Paradigms in Machine Translation.}  
Alignment techniques in neural machine translation have evolved from Minimum Risk Training~\citep{shen2015minimum} to more sophisticated reinforcement learning approaches~\citep{dang-etal-2024-rlhf}. While PPO-based RLHF has achieved success in general-domain alignment, its application to MT presents unique challenges, particularly due to the need for fine-grained quality signals rather than the bandit reward. Recent works like \citet{he2024improving} and \citet{xu2024contrastive} have investigated the use of automatic metrics to select better translations or construct preference pairs to improve the LLM, while \citet{marco-o1} explored scaling test-time compute to further enhance translation performance. Recently, \citet{ramos2024fine} pioneered the use of xCOMET as a dynamic reward signal during RL training. However, these methods remain limited to sequence-level guidance or binary approximations of the reward process, failing to provide the fine-grained token-level feedback required for more accurate translation alignment.

\section{Conclusion}
In this work, we propose MT-RewardTree, a comprehensive framework for constructing, evaluating, and deploying process reward models in MT. Our framework leverages an automatic token-level preference pair generation approach inspired by approximate Monte Carlo Tree Search, effectively addressing the challenge of large-scale fine-grained supervision annotation. Extensive experiments on both sequence-level and token-level benchmarks demonstrate that our MT-PRM achieves advanced performance in reward modeling in MT, surpassing traditional sequence-level preference pairs. Our exploration of token credit assignment and test-time alignment provide valuable insights for the application of reward models in MT.

\section*{Limitations}
Although we have developed the first comprehensive framework for process reward models in the field of machine translation, several important challenges remain to be addressed. Our work primarily focuses on synthesizing token-level data to leverage its fine-grained benefits. However, methods like Token-level DPO, RTO which optimize training algorithms, also show promise in further improving PRM performance. Additionally, our current framework includes only a limited set of high-resource languages, and expanding to multilingual settings, especially for low-resource languages, is a crucial direction for future work. While we have demonstrated the potential applications of reward models in test-time alignment and hypothesis ensembling, their integration into reinforcement learning training remains an important area for exploration.

\section*{Acknowledgments}
This work is supported by the National Key R\&D Program of China (Grant No. 2024YFC3308304), the "Pioneer" and "Leading Goose" R\&D Program of Zhejiang (Grant no. 2025C01128), 
the National Natural Science Foundation of China (Grant No. 62476241), the Natural Science Foundation of Zhejiang Province, China (Grant No. LZ23F020008), the Zhejiang Zhejiang Key Laboratory of Medical Imaging Artificial Intelligence.

\bibliography{custom}

\appendix

\end{document}